\documentclass[review]{elsarticle}
\usepackage[colorlinks=true,linkcolor=blue]{hyperref}
\usepackage{lineno}
\usepackage{multirow}
\modulolinenumbers[1]
\journal{NeuroImage}

\biboptions{semicolon,round,sort,authoryear}

\begin{document}
\begin{frontmatter}
\title{SHARM: Segmented Head Anatomical Reference Models}
\author[a]{Essam A. Rashed}
\ead{rashed@gsis.u-hyogo.ac.jp}
\author[b]{Mohammad Al-Shatouri} 
\author[c,d]{Ilkka Laakso} 
\author[e,f]{Akimasa Hirata} 
\address[a]{Graduate School of Information Science, University of Hyogo, Kobe 650-0047, Japan}
\address[b]{Radiology Department, Faculty of Medicine, Suez Canal University, Ismailia 41522, Egypt}
\address[c] {Department of Electrical Engineering and Automation, Aalto University, Espoo, Finland}
\address[d] {Aalto neuroimaging, Aalto University, Espoo, Finland}
\address[e]{Department of Electrical and Mechanical Engineering, Nagoya Institute of Technology, Nagoya 466-8555, Japan}
\address[f]{Center of Biomedical Physics and Information Technology, Nagoya Institute of Technology, Nagoya 466-8555, Japan}

\begin{abstract}

Reliable segmentation of anatomical tissues of human head is a major step in several clinical applications such as brain mapping, surgery planning and associated computational simulation studies. Segmentation is based on identifying different anatomical structures through labeling different tissues through medical imaging modalities. The segmentation of brain structures is commonly feasible with several remarkable contributions mainly for medical perspective; however, non-brain tissues are of less interest due to anatomical complexity and difficulties to be observed using standard medical imaging protocols. The lack of whole head segmentation methods and unavailability of large human head segmented datasets limiting the variability studies, especially in the computational evaluation of electrical brain stimulation (neuromodulation), human protection from electromagnetic field, and electroencephalography where non-brain tissues are of great importance. 

To fill this gap, this study provides an open-access Segmented Head Anatomical Reference Models (SHARM) that consists of 196 subjects. These models are segmented into 15 different tissues; skin, fat, muscle, skull cancellous bone, skull cortical bone, brain white matter, brain gray matter, cerebellum white matter, cerebellum gray matter, cerebrospinal fluid, dura, vitreous humor, lens, mucous tissue and blood vessels. The segmented head models are generated using open-access IXI MRI dataset through convolutional neural network structure named ForkNet$^+$. Results indicate a high consistency in statistical characteristics of different tissue distribution in age scale with real measurements. SHARM is expected to be a useful benchmark not only for electromagnetic dosimetry studies but also for different human head segmentation applications.

\end{abstract}

\begin{keyword}
Human head models, brain segmentation, convolutional neural networks, MRI
\end{keyword}

\end{frontmatter}

\section{Introduction}

Anatomical reference models of human subjects are of great importance in several computer simulation studies such as medical imaging, dosimetric evaluation for diagnosis and therapy and human safety. In principle, digital models are generated from anatomical imaging of real subjects for better understanding real physical effects. Specifically, personalized electrode positions or coil location are explored in the non-invasive electrical and magnetic stimulation~\citep{ANTONENKO20191159}, in addition to group-level optimization\citep{Laakso2015,Gomez2018}. Variability analysis is needed to derive the limit in human protection from electromagnetic field~\citep{Hirata2021TEMC,international2020gaps}.

Several attempts provided different models that represent whole body models~\citep{Nagaoka2004PMB,Kim2008PMB,Christ2009PMB,Segars2010MP,Yu2015PMB}. A useful review is in ~\citep{Kainz2019TRPM}. Segmentation of brain tissues is of high interest in several clinical applications such as diagnosis of abnormalities, assessment of neurophysiological performance, surgery planing and many others. Most of standard medical imaging applications can represent brain tissues in high contrast which enable accurate automatic annotation~\citep{Baur2021MIA}. However, segmentation of non-brain tissues is challenging as it represented in low contrast and/or allocated in limited regions. Moreover, in clinical medical applications, imaging protocols are usually adjusted such that brain tissues are presented in high quality as the main target of diagnostic applications~\citep{kalavathi2016methods}.

Whole head segmentation have been discussed mainly for the development of digital models for electromagnetic stimulation studies. SimNibs is an open source software for the simulation of non-invasive brain stimulation that include magnetic resonance (MR) image segmentation to generate head models \citep{simnibs,Saturnino500314}. However, segmentation is limited to major head tissues such as white matter (WM), grey matter (GM), cerebrospinal fluid (CSF), skull and scalp. ROAST is another pipeline the include automatic MRI segmentation based on SPM12~\citep{ASHBURNER2005839} with variety of segmentation and electromagnetic modeling options~\citep{Huang2019}. However, ROAST segmentation is also limited to a few number of tissues as in Table~1 in Ref.~\citep{Huang2019}. Segmentation of fifteen head tissues using multi-modality images (MRI T1/T2, mDixon, venogram and CT) is proposed in~\citep{Puonti2020}. Recently, the use of deep learning architectures demonstrate quality improvement of anatomical segmentation~\citep{akkus2017deep}. Several network architectures such as ForkNet~\citep{Rashed2019NI}, SubForkNet~\citep{Rashed2020NN} and FastSurfer~\citep{Henschel2020NI} have been used to generate human head models with different scope and applications.

Due to the complexity of full head segmentation and requirements of intensive efforts for manual parameters adjustment, there is a shortage of relatively large dataset of human head models. This problem becomes more feasible with the use of deep learning as robust segmentation tool with superior accuracy compared to conventional methods. The aim of this work is to generate an open-access Segmented Head Anatomical Reference Models (SHARM) that is large enough for subject variability studies. The developed dataset consists of 196 subjects segmented into 15 different tissues. The main contributions of this study can be summarized as follows:

\begin{itemize}
\item An open-source deep learning pipeline for automatic segmentation of MRI head images.
\item An open-access large human head dataset segmented into brain and non-brain tissues.
\item Evaluation of the consistency of segmented models with realistic tissue characteristics.
\end{itemize}

\section{Materials and methods}

\subsection{Dataset and general pipeline}

The MRI dataset used in this study is the IXI Dataset\footnote{\href{http://brain-development.org/ixi-dataset/}{http://brain-development.org/ixi-dataset/}} which consists of around 600 MRI scans of healthy subjects. A set of 196 subjects are selected (123 females, 70 males, and 3 unknown), that are imaged at two hospitals (100 were imaged at the Guy's Hospital (London, UK) with Philips 1.5T system and 96 were images at the Hammersmith Hospital (London, UK) using a Philips 3T system). Excluded images criteria are based on quality of the image and availability of multi-modlity scans. The T1w/T2w image data are in Nifti formats and are used for generation of the head models.

\begin{figure*}
\centering
\includegraphics[width=1.1\textwidth]{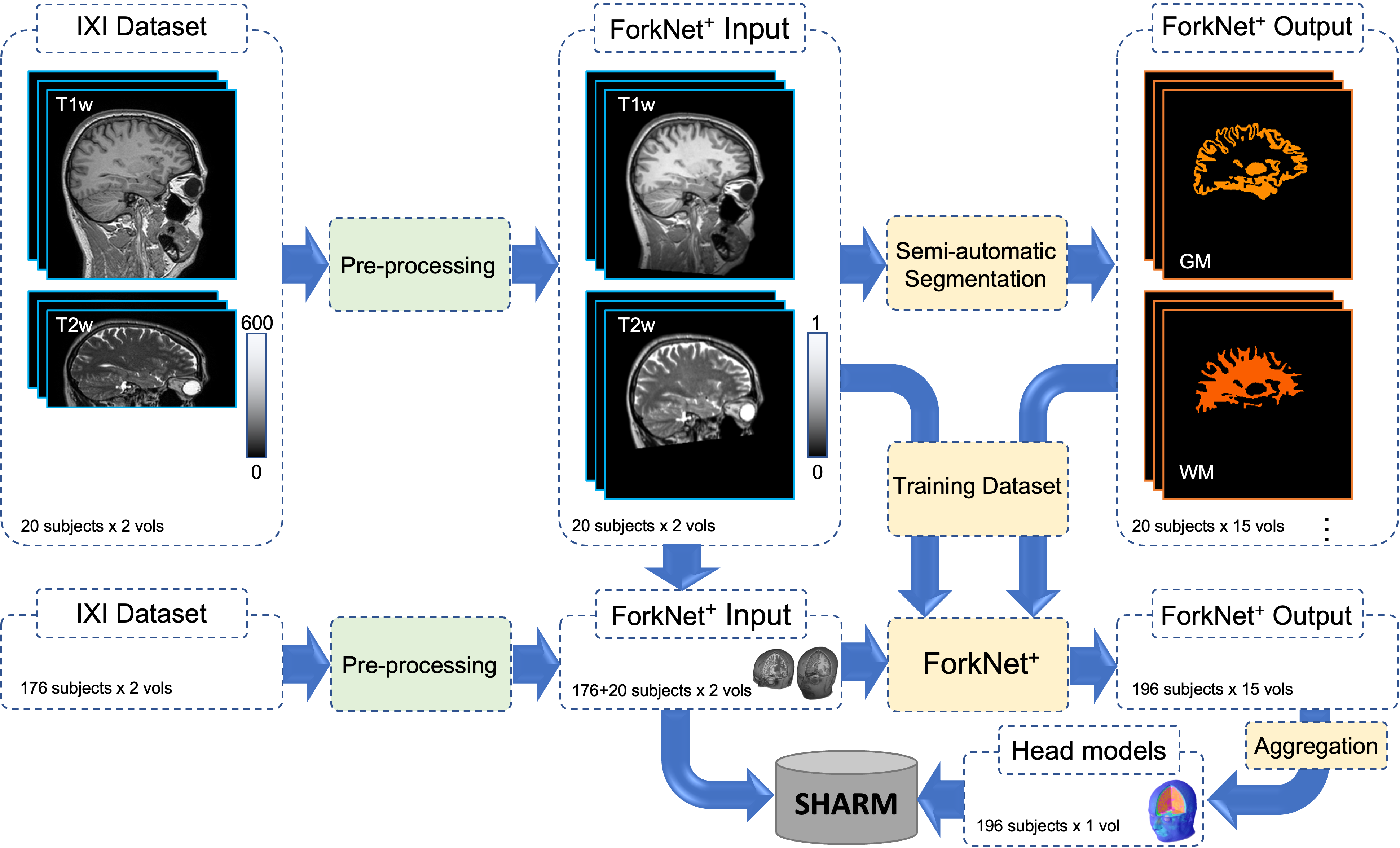}
\caption{Data flow used to generate SHARM from IXI dataset.} 
\label{pre1}
\end{figure*}

The raw T1 and T2-weighted MR images are registered using non-rigid registration such that T2w are adjusted to fit with T1w. The acoustic noise is reduced through contouring of head surface and relabeling external region as air voxels. The N4 bias field correction method (ITK\footnote{\href{https://itk.org/}{https://itk.org/}}) is used for bias correction of both MRI modalities. Both T1w/T1w images are normalized with zero mean and unit variance, then scaled to vales [0.01, 0.99]. All the above pre-processing procedures are used to generate the network input volumes (a set of two 256$^3$ volumes representing T1w/T2w MRI with unified 1$^3$mm resolution). A selected set of the network input is segmented into 15 different head tissues using the semi-automatic method detailed in~\citep{Laakso2015}. The segmentation binary labels are used as network target (output) through training process. The remaining subjects are evaluated using trained network to automatically generate segmentation labels. Finally, an aggregation process is used to combine different segmentation labels into a head model. The pre-processed MR scans and segmented head models are available for each subject in SHARM dataset. The data processing pipeline is shown in Fig.~\ref{pre1}.

\begin{figure*}
\centering
\includegraphics[width=1.0\textwidth]{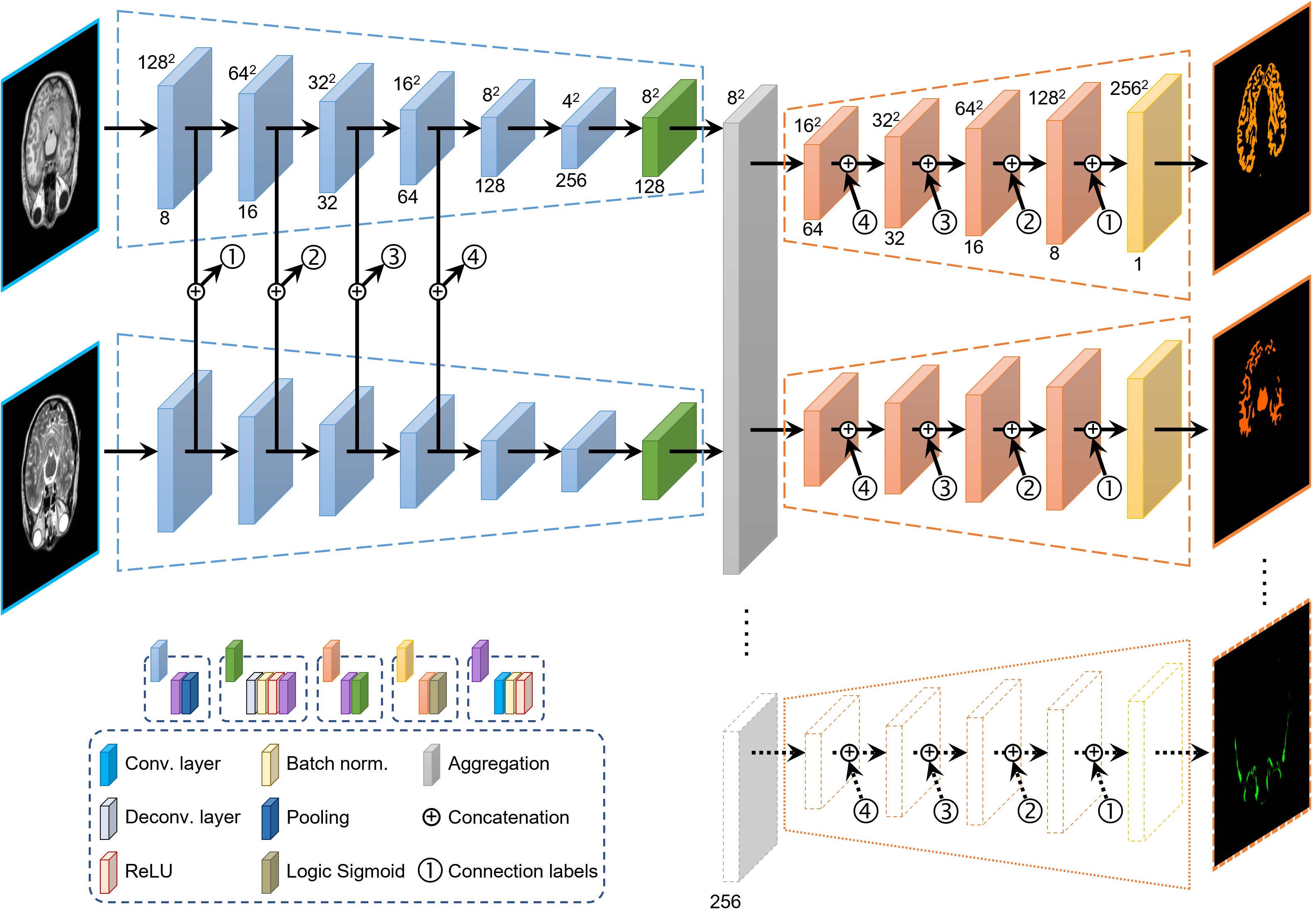}
\caption{ForkNet$^+$ with MRI T1w/T2w inputs and $N$ segmented tissues outputs.} 
\label{network}
\end{figure*}

\subsection{Semi-automatic segmentation}

The target dataset for the training process was generated using a semi-automatic segmentation pipeline that segments T1- and T2-weighted MR image data into 15 tissue types \citep{Laakso2015}. Briefly, after bias correction and normalization of the MR data, the pipeline first splits the head into three compartments: inner compartment, consisting of the volume inside the inner surface of the skull; middle compartment, consisting of the skull and nasal cavity; and the outer compartment, consisting of the volume between the outer surface of the skull and the outer surface of the skin. The quality of these compartments is verified by visual inspection, and whenever necessary, control parameters are manually altered until the compartments match the MR data.

The inner compartment is segmented into brain using FreeSurfer image analysis software \citep{DALE1999179,fischl2000measuring}. The brain segmentation consists of cerebral gray matter, cerebral white matter, cerebellar gray matter, cerebellar white matter, deep brain structures (brainstem, accumbens, amygdala, caudate, hippocampus, pallidum, putamen, thalamus), and ventricular CSF. The remaining non-brain volume in the inner compartment is segmented into CSF (bright T2-weighted image), blood (dark T2), and dura (non-brain non-CSF tissue close to the inner boundary of the skull). Anterior and middle cerebral arteries initially estimated from T2 are corrected using thresholding of registered MRA images (when available). Deep brain structures are treated as GM. The middle compartment consisting of the skull is segmented into cortical and cancellous bone by thresholding the T2-weighted MRI data. It is ensured that the inner and outer cortical bone layers are at least 1~mm and 1.5~mm thick, respectively. The nasal cavity also belongs to the middle compartment and is segmented as either mucous tissue or cortical bone based on T2-weighted images. The outer compartment is segmented into skin, fat, muscle, and eyes. The scalp (including subcutaneous fat) is segmented as the outer layer of the head, with thickness between 2~mm and 10~mm. Fat and muscle are segmented based on thresholding the T1-weighted image data. Finally, eyes and lens are segmented using both T1- and T2-weighted image data.

The resulting segmentation has uniform voxel size of 0.5~mm$\times$0.5~mm$\times$0.5~mm, half of that of the input MR images. In this study, the segmented dataset was downscaled to the same resolution as the input images using three-dimensional nearest neighborhood interpolation algorithm.

\subsection{Network architecture}
The deep learning architecture used here is an extension of ForkNet \citep{Rashed2019NI} by considering input data from both T1- and T2-weighted MRI scans. We then refer to the new network as ForkNet$^+$. The network inputs are the MRI scans in two encoders and outputs are $N$ decoders each assigned to single anatomical structure (here $N$=15). The details of the layer structures and data processing flow is shown in Fig.~\ref{network}. The network output are binary masks that identify different anatomical tissues/liquids such as skin, muscle, fat, skull (cortical bone), skull (cancellous bone), CSF, blood vessels, dura, brain GM, brain WM, cerebellum GM, cerebellum WM, vitreous humor, eye lens, mocous tissue and whole head. ForkNet$^+$ design is flexible and easy to be adjusted to segmented specific number of tissues for different applications as each tissue is segmented using separate decoder (Fig.~\ref{network}).

\begin{figure*}
\centering
\includegraphics[width=1.1\textwidth]{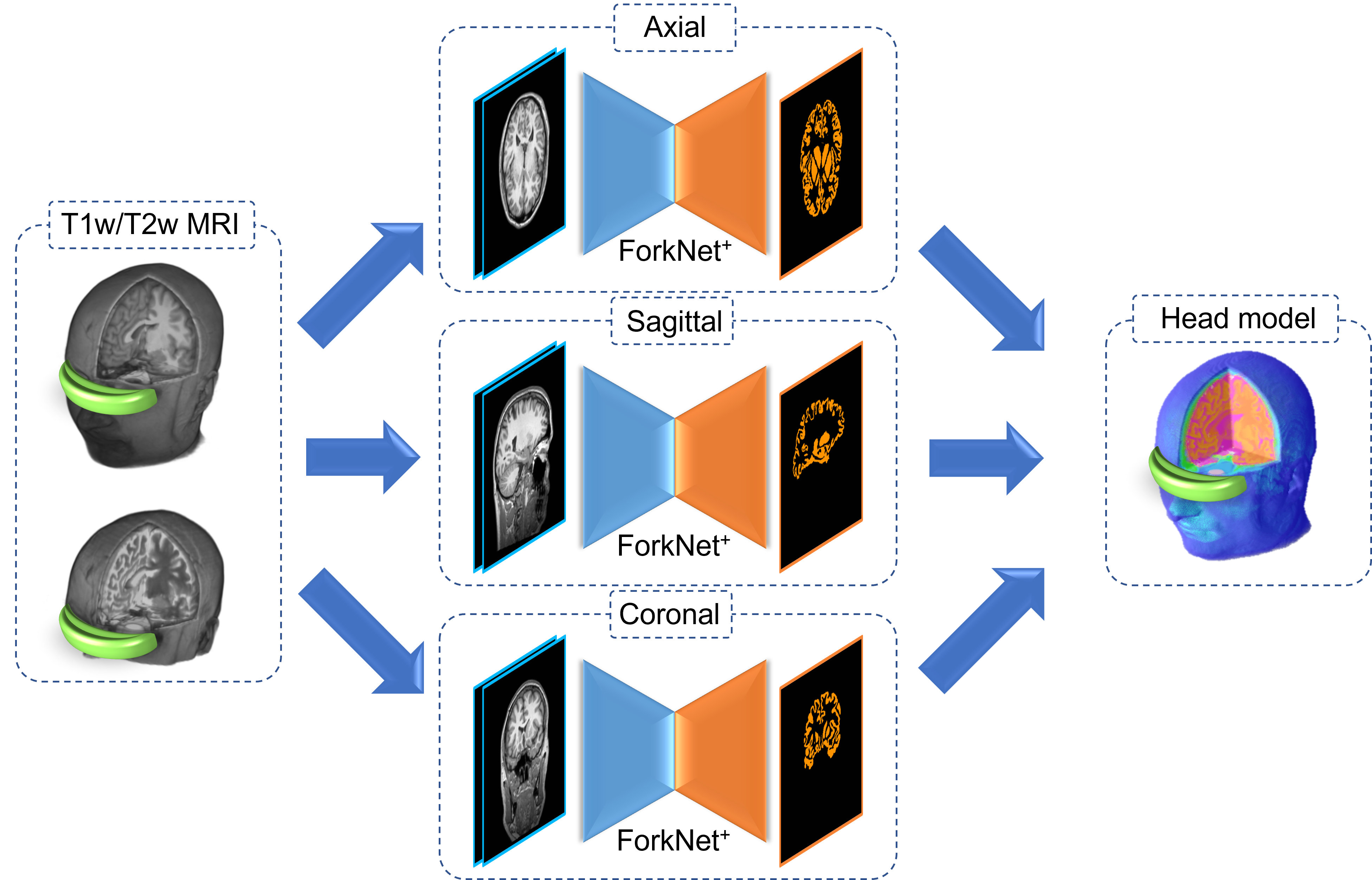}
\caption{Network evaluation through different direction to generate the head model.}
\label{model1}
\end{figure*}

\subsection{Head model generation}

Once the network is well-trained, the head models are generated through fast evaluation process. To reduce artifacts caused by 2D slice segmentation, a set of three networks are trained using slices of axial, sagittal and coronal directions as shown in Fig.~\ref{model1}. A rule-based segmentation merge approach using majority vote is used to generate the final segmentation from different slicing directions. When no majority in a voxel is found, the neighborhood majority vote~\citep{Rashed2020NN}.

\begin{figure*}
\centering
\includegraphics[width=1.0\textwidth]{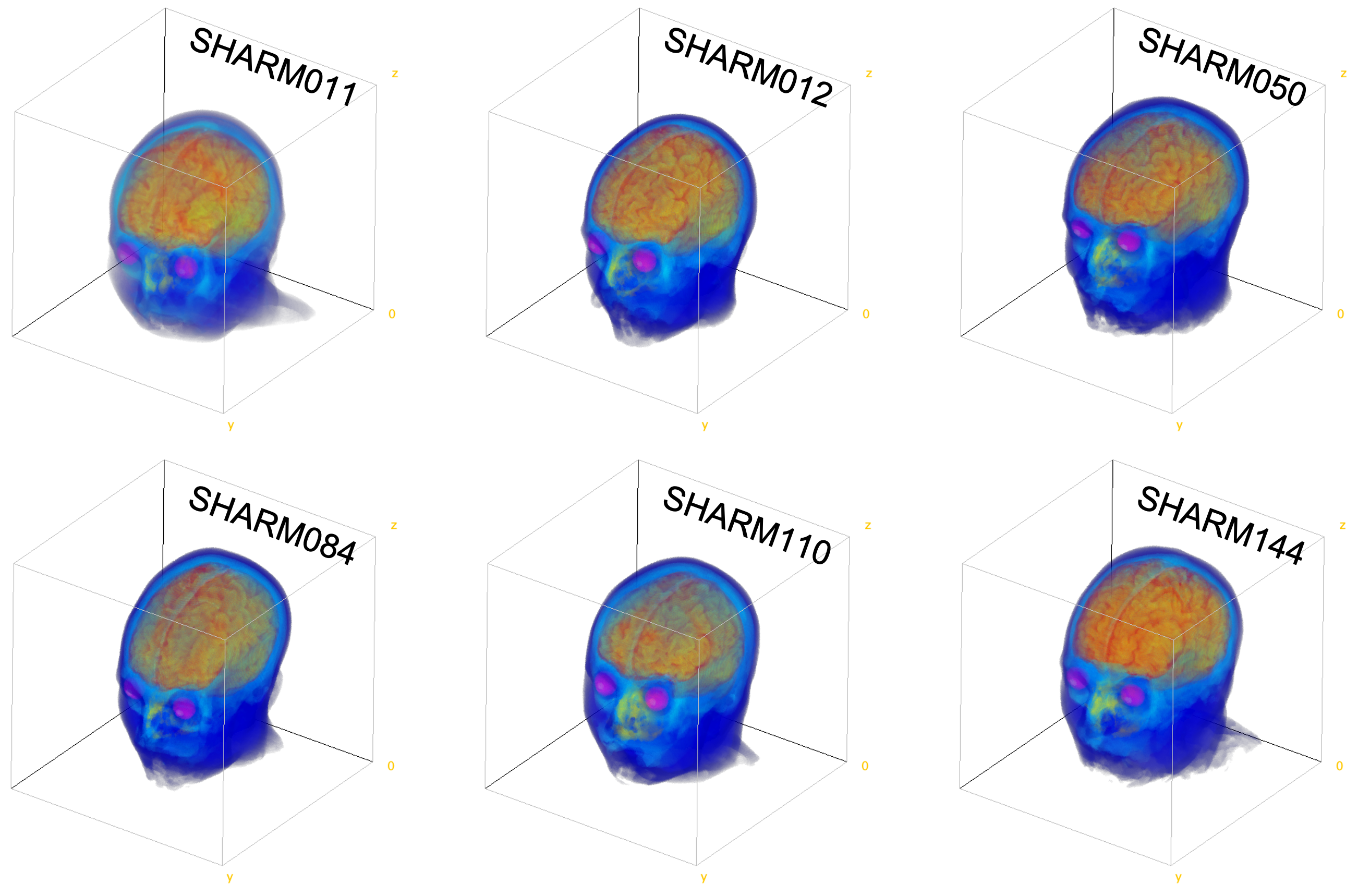}
\caption{Volume rendering sample of generated head models.} %
\label{sample_1_6}
\end{figure*}

\begin{figure*}
\centering
\includegraphics[width=1.1\textwidth]{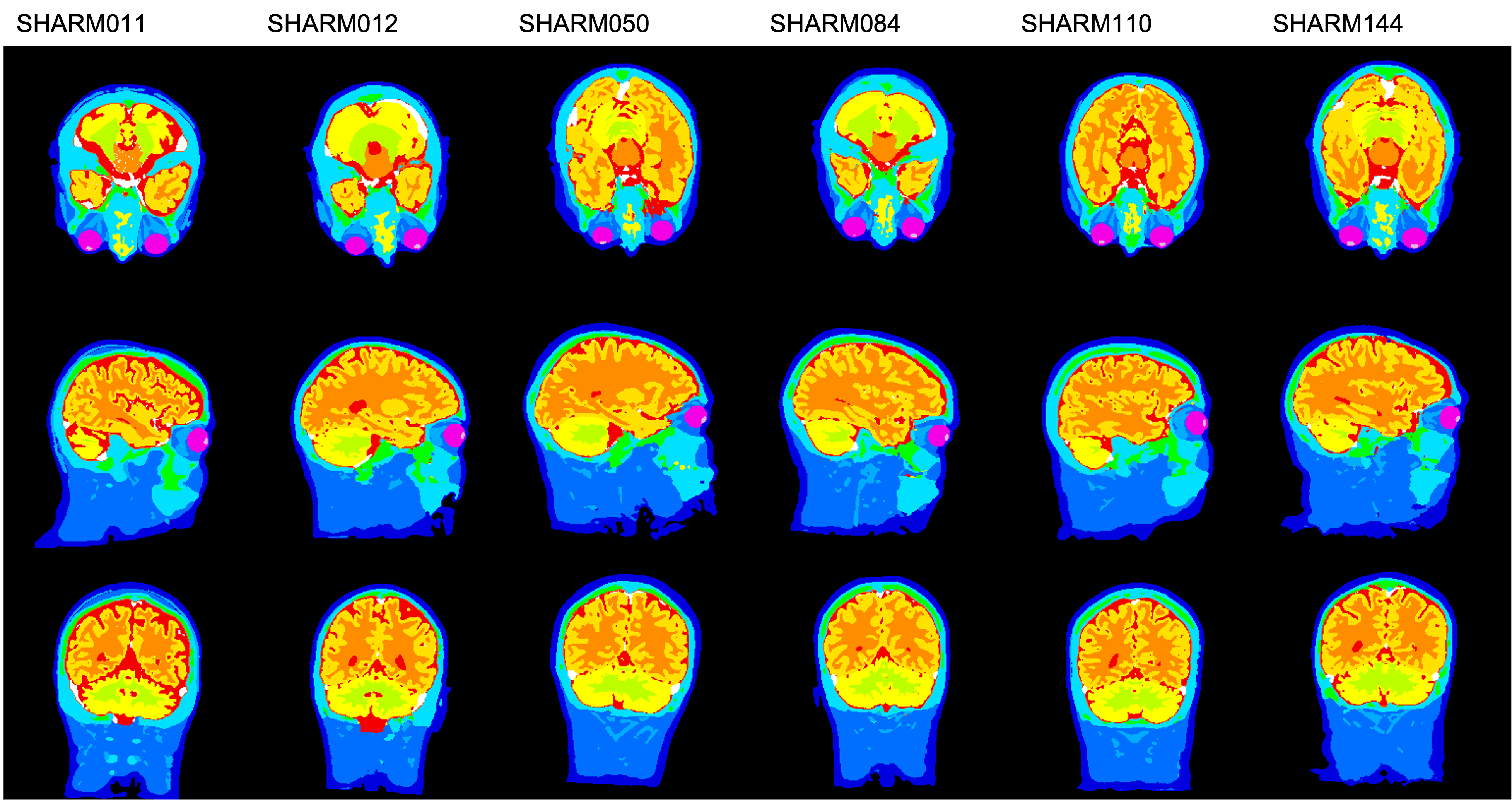}
\caption{Axial, sagittal and coronal slices (top to bottom) of head models shown in Fig.~\ref{sample_1_6} in order.}
\label{sample_slices}
\end{figure*}

\section{Results}

A set of 20 randomly selected head models and associated segmented labels are used to train ForkNet$^+$. The network architecture is developed using Wolfram Mathematica (R) ver. 13.0, installed on a Ubuntu 20.04 workstation of 12 Cores Intel (R) Core (TM) i9-10920X @3.50GHz, 64 GB memory, and NVIDIA RTX A6000 GPU. A three networks (axial, sagittal and coronal) are trained with cross-entropy loss function and ADAM optimization algorithm. The training was considered using 50 epochs with batch size 4. To reduce the computation cost, the number of output tracks is set to $N=4$ (i.e., a set of 4 tissues are trained simultaneously). A single training round requires about 24 mins.

The remaining 176 head models are evaluated through trained networks and the network output is aggregated to generate the head models. Example of generated head models are shown in Figs.~\ref{sample_1_6} and~\ref{sample_slices}. In some few cases, some manual edition is required mainly to remove a small regions of CSF-like tissue uncorrected segmented inside the mouth. Some other cases are excluded due to the strong noise that is difficult to be automatically removed and lead to incorrect segmentation of external contour. Evaluation of segmentation accuracy is not conducted due to the lack of manual true annotation and it is out of the scope of this work. However, we provide quantitative assessment of different tissues of the SHARM dataset. In principle, we study the variability of segmented volumes within age scale to validate the validity of SHARM models. 

\begin{figure*}
\centering
\includegraphics[width=1.2\textwidth]{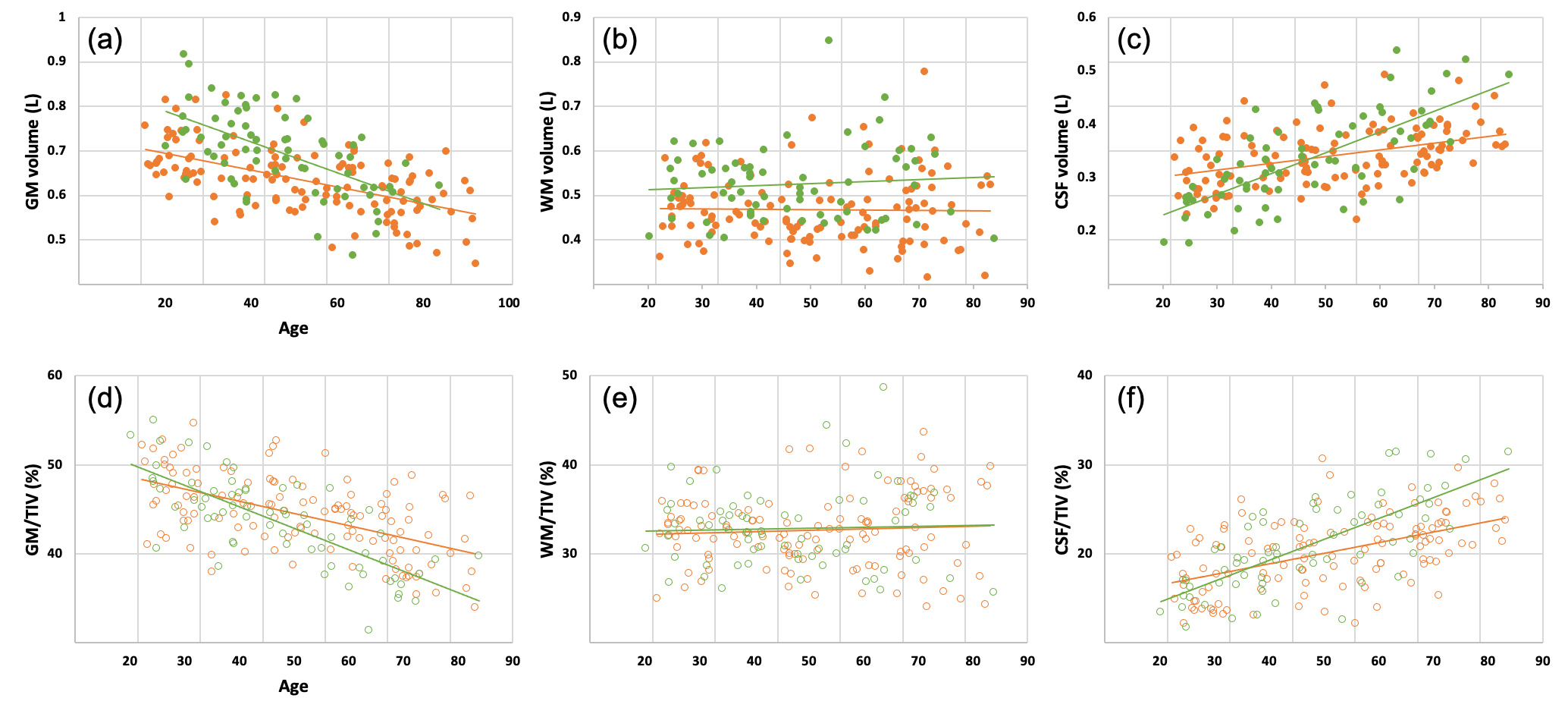}
\caption{(a) Regression lines with scatter plot of total GM volume over age for all subjects (female in orange and male in green). (b) and (c) are regression lines and scatter plots for total WM and CSF, respectively. (d)-(f) are the regression lines and scatter plots for fractional volume (with respect to TIV) of GM, WM and CSF, respectively.}
\label{data1}
\end{figure*}

\begin{figure*}
\centering
\includegraphics[width=.8\textwidth]{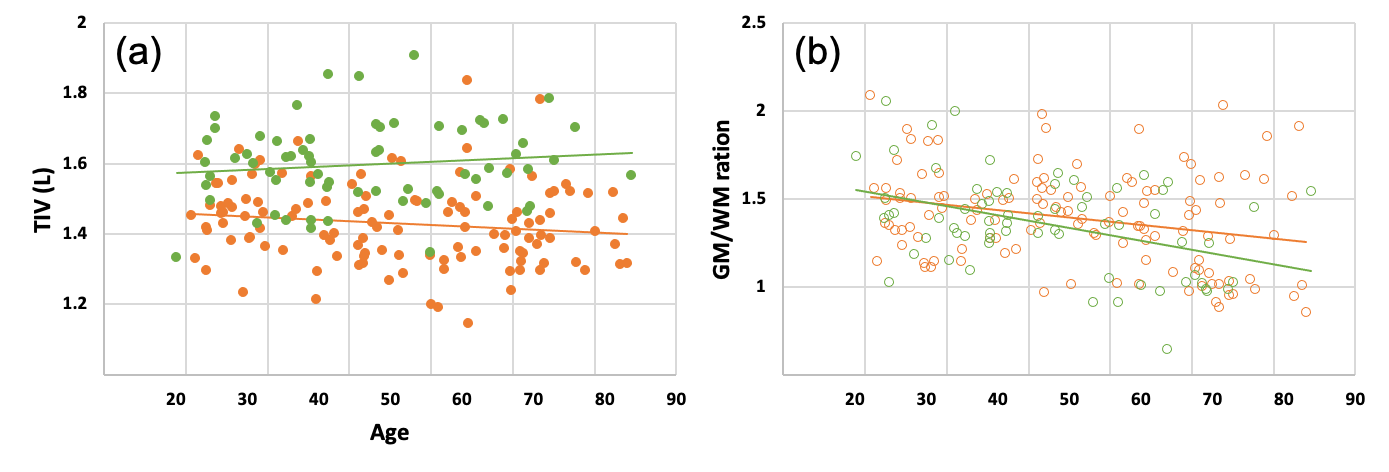}
\caption{(a) Regression curves and scatter plot of (a) total intracranial volume (TIV) in liters and (b) GM/WM ratio over age for all subjects.}
\label{data3}
\end{figure*}

\begin{figure*}
\centering
\includegraphics[width=1.2\textwidth]{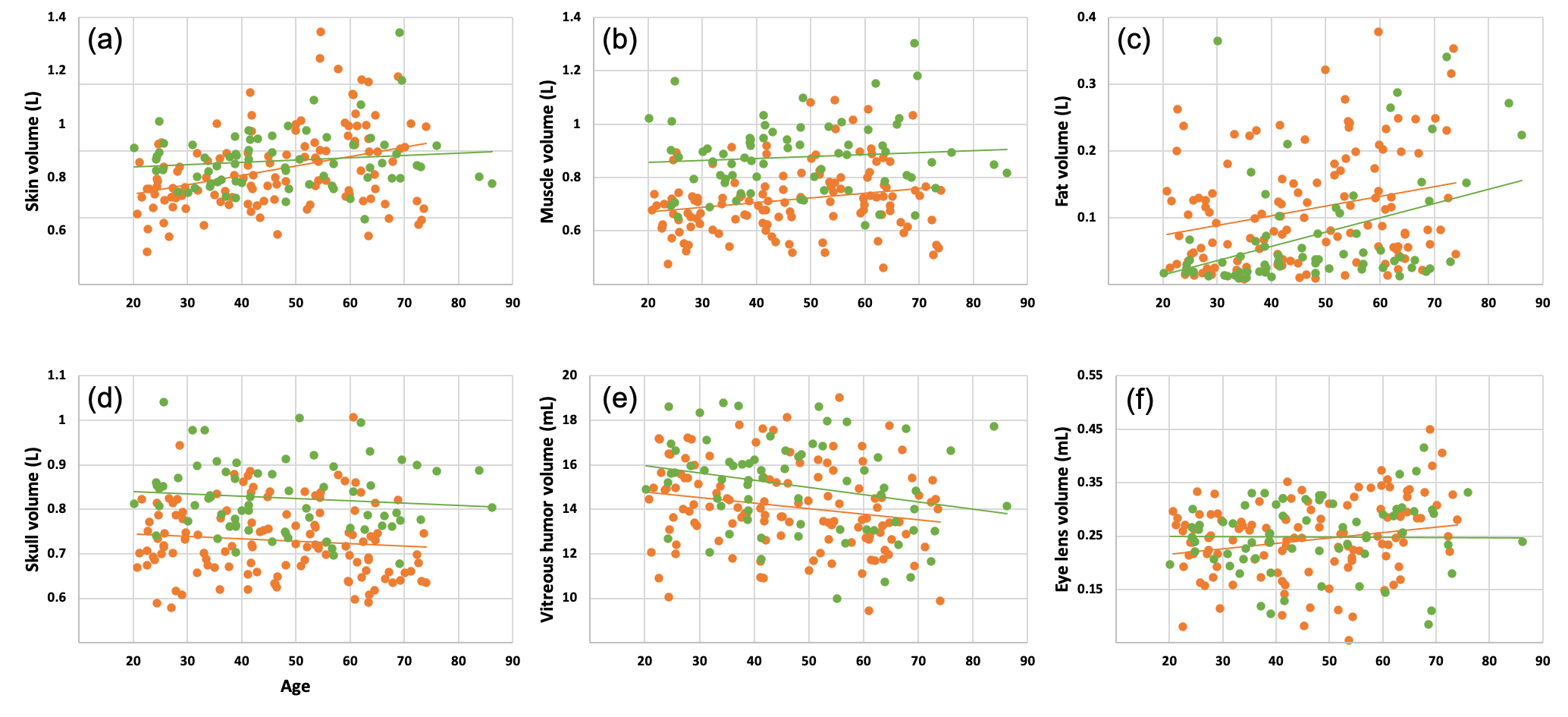}
\caption{Regression curves of segmented volume of (a) skin, (b) muscle,  (c) fat, (d) skull, (e) vitreous humor, and (f) eye lens in SHARM models.} %
\label{data2}
\end{figure*}

\begin{figure*}
\centering
\includegraphics[width=.4\textwidth]{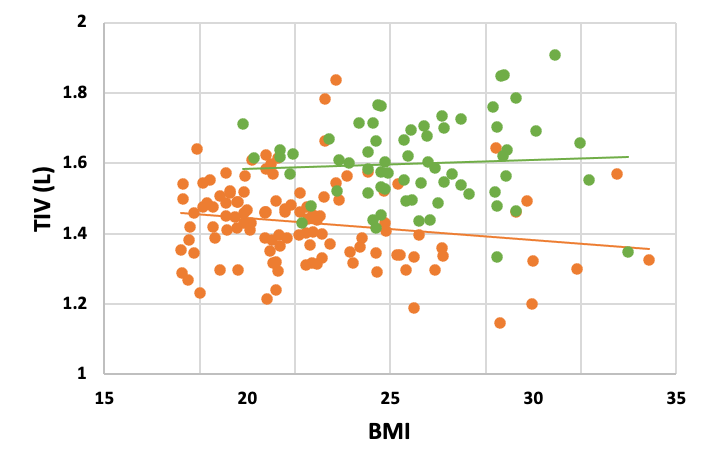}
\caption{Regression curves of TIV per BMI.} %
\label{data4}
\end{figure*}

\begin{figure*}
\centering
\includegraphics[width=0.5\textwidth]{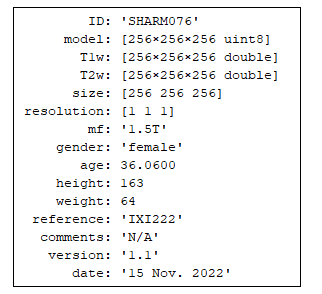}
\caption{Structure of subject SHARM076 that include MRI scans, head model and other subject details.} %
\label{structure}
\end{figure*}

Figure~\ref{data1} demonstrates a regression curves of segmented brains. Brain is considered as a composition of GM, WM and CSF. It shows that a decline in global GM volume with age ($R^2$=0.274), while there is no significant change in WM volume with age ($R^2$=0.001). A remarkable increase in CSF volume with age ($R^2$=0.277). Difference between gender grouping and percentage of different structures with respect to total intracranial volume (TIV) are also shown in Fig.~\ref{data1} and it is consistent with  results reported in literature (e.g.,~\citep{Ruigrok2014NBR, Ge2002AJN, GOOD200121}). The change in TIV over age and the correlation between GM/WM ratio with age is shown in Fig.~\ref{data3}. The change of skin, muscle, fat, skull, vitreous humor, and eye lens volumes is shown in Fig.~\ref{data2}. In general the volume of skin, muscles, and fat tissues are increasing with aging.  Skull and eye lens does not change so much while the volume of vitreous humor is shrinked as reported earlier~\citep{Sebag1987N}. Data shown in Fig.~\ref{data4} demonstrate that the volume of brain is highly correlated with the body mass index (BMI) and brain volume of male subjects are of large volume compared with females \citep{Eliot2021NBR}. SHARM dataset can be download\footnote{\href{https://figshare.com/s/a4d9ba6f18a6b7f7ba2c}{https://figshare.com/s/a4d9ba6f18a6b7f7ba2c} (zip, 7.62 GB)} in MATLAB (*.mat) files with structure shown in Fig.~\ref{structure}.


\section{Discussion}

\begin{figure*}
\centering
\includegraphics[width=1.0\textwidth]{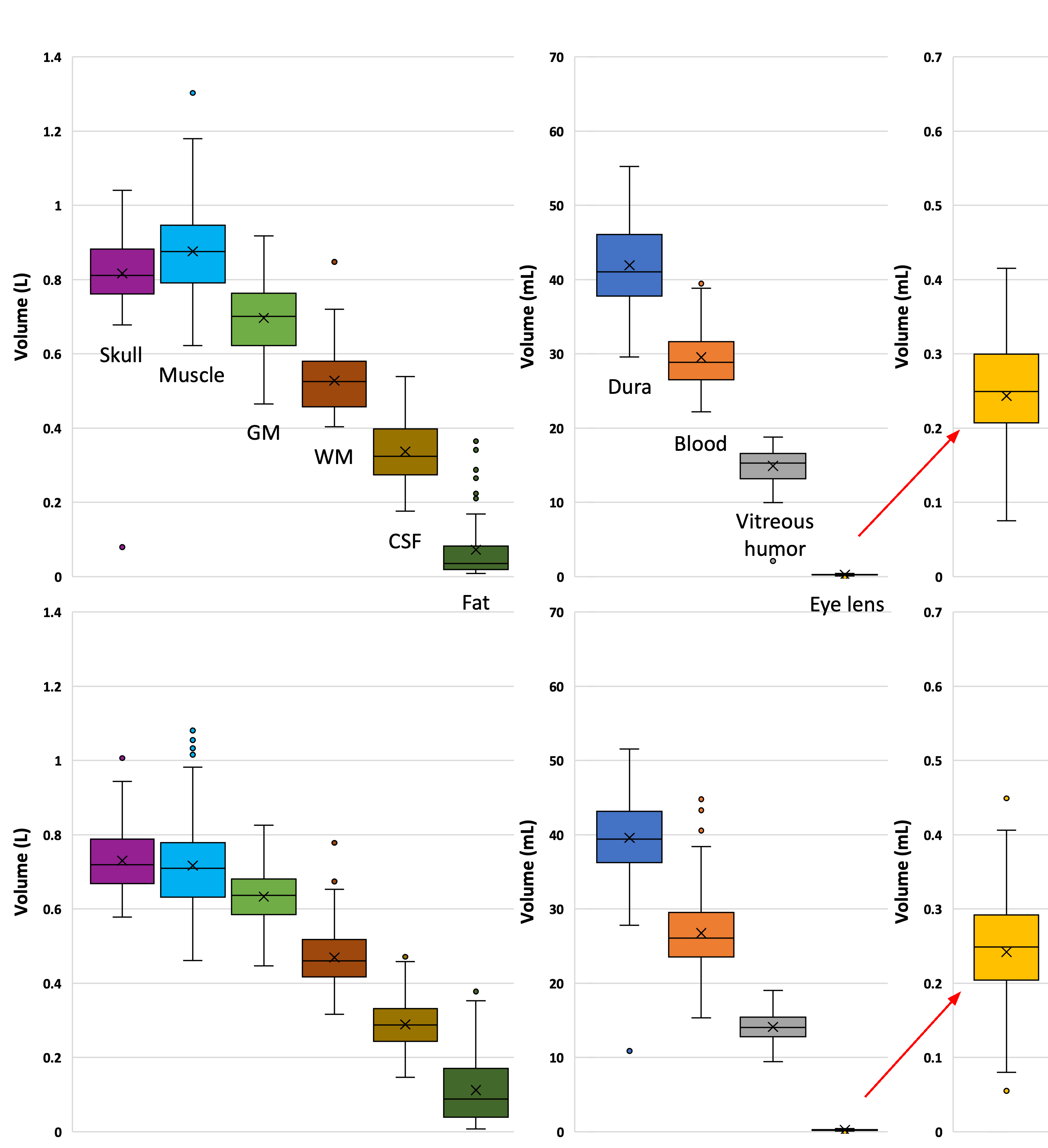}
\caption{Boxplot demonstrates volume variations of different structures in SHARM male (top) and female (bottom) subjects.} %
\label{boxplot}
\end{figure*}

This work demonstrate a new benchmark dataset for several computational neuroscience applications. The open-access SHARM consists of 196 head models segmented into 15 different tissues that cover a wide range of subject variability. A boxplot demonstrates the volume variations of different structures in SHRAM is shown in Fig.~\ref{boxplot}. It is clearly observed that skull volume is 0.827 $\pm$ 0.08 $L$ (male) and 0.730 $\pm$ 0.08 $L$ (female). These values are highly correlated with those listed in the ICRP Reference man (averaged bone without marrow of Skull is 708 gm). The vitreous humor is 15.098 $\pm$ 2.10 $mL$ (male) and 14.124 $\pm$ 1.91 $mL$ (female) that is referenced as value 15 $\pm$ 6.5 gm in adult~\citep[Table 97, p. 220]{refman}. Also, eye lens is 0.246 $\pm$ 0.07 mL (male) and  0.242 $\pm$ 0.08 mL (female) which is calculated as 172 to 258.1 gm for (20 - 60 y) adult~\citep[Table 100, p. 225]{refman}. Calculated volume and weight of different SHRAM tissues are compared with those reported in~\citep{refman} in Table~\ref{Tab1}. These values are example that indicate segmentation accuracy of non-brain tissues. 

\begin{table*}
\centering
\caption{Comparison of volume/weight values of SHARM models with ICRP Reference man~\citep{refman}.}
\label{Tab1}
\includegraphics[width=1.0\textwidth]{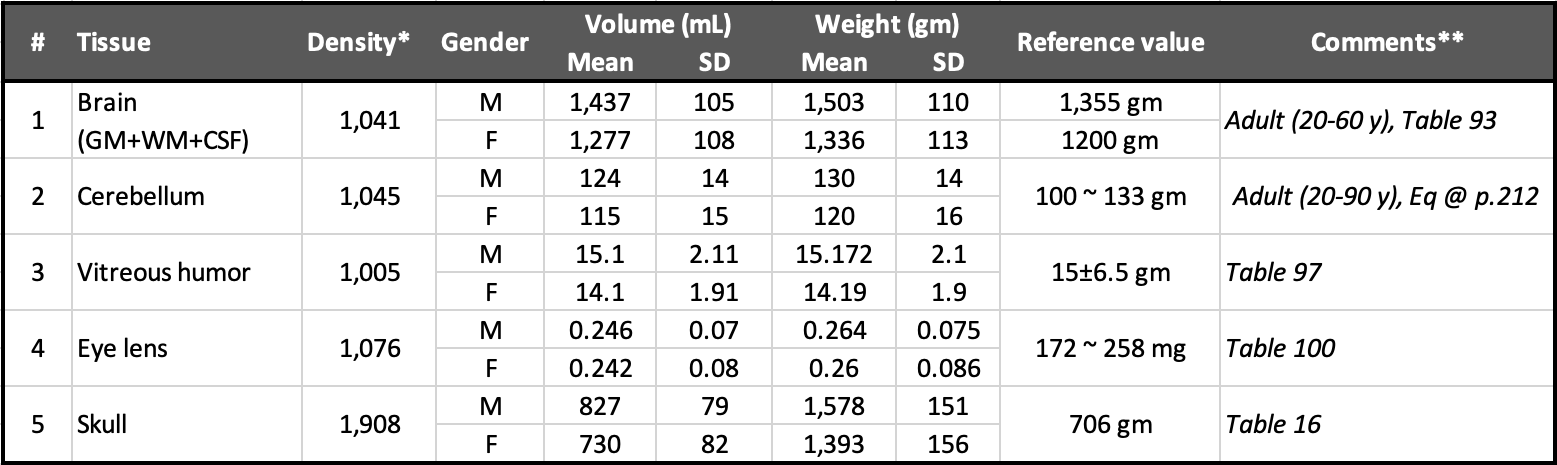}
\begin{tabular}{l}
\emph{$^*$Average density values are acquired from~\citep{den}}\\
\emph{$^{**}$References from~\citep{refman}}\\ 
\end{tabular}
\end{table*}

The segmented models along with normalized T1- and T2-weighted MR scans are available for each subject in additional to other demographic information. Most of the models in SHARM are presented with full head and neck segmentation which enables simulation studies that requires full head models. moreover, the trained deep learning model used to generate SHARM is shared which can be used to generate additional models considering the availability of consistent MR scans. The software ForkNet$^+$ generate individual tissue segmentation in terms of probability maps which enables customized segmentation of a single subject through weighting based aggregation process (similar to those presented in~\citep{Rashed2021PMB}). It is worth noting that evaluation of segmentation accuracy is out of the scope of this work, because earlier version of the ForkNet~\citep{Rashed2019NI} segmentation have been evaluated. 

The limitation of this work is the lack of variability of MR data acquisition. Data are acquired from two scanners installed at two medical institutes but it is developed by the same manufacturer.  Further extension with data from other manufacturers are planned to be included in future versions. Moreover, we will include more information considering the segmentation of deep brain structures and fiber orientations in future versions of SHARM. Also, there is a lack of bone structure accuracy in the neck region as lack of neck data in T2 images. In future, we will investigate potential approaches to improve the segmentation neck region properly.


\section{Conclusion}

In this study, we present SHARM, a benchmark dataset of 196 segmented human head models. The models are segmented into 15 different tissues using deep learning network named ForkNet$^+$. The freely available models along with normalized MR T1- and T2-weighted scans would enable a large scale studies in different applications such as  electromagnetic brain stimulations. Results demonstrate that the segmented models are of high consistency with measurements obtained from real measurements. One feature of ForkNet$^+$ is the segmentation of each tissue is generated as probability maps that enable parametric segmentation for further customization of the generated head models. The trained networks as well as source code are shared for potential usage of head models generation. With large scale of subject age, SHARM would enable different electromagnetic dosimetry and human safety studies in a reliable manner. After publication, Mathematica notebooks demonstrate the implementation of ForkNet$^+$ architectures and trained networks will be available for download at: \href{http://github.com/erashed/ForkNetPlus}{https://github.com/erashed/ForkNetPlus}


\section*{Acknowledgment}

This work was funded by the Japan Society for the Promotion of Science (JSPS), a Grant-in-Aid for Scientific Research, Grant number JSPS KAKENHI 22K12765.

\bibliography{Refs}

\end{document}